\documentclass[letterpaper]{article} 
\usepackage{aaai24}  
\usepackage{times}  
\usepackage{helvet}  
\usepackage{courier}  
\usepackage[hyphens]{url}  
\usepackage{graphicx} 
\usepackage{natbib}  
\usepackage{caption} 
\usepackage{algorithm}
\usepackage{algorithmic}
\usepackage{booktabs}
\usepackage{makecell}
\usepackage{multirow}
\usepackage{multicol}
\usepackage{graphicx}
\usepackage{xcolor}
\usepackage{amssymb} 
\newcommand{\myparagraph}[1]{\vspace{2pt}\noindent{\bf{#1}}~~}

\usepackage{newfloat}
\usepackage{listings}
\DeclareCaptionStyle{ruled}{labelfont=normalfont,labelsep=colon,strut=off} 
\floatstyle{ruled}
\newfloat{listing}{tb}{lst}{}
\floatname{listing}{Listing}

\title{LAMM: Label Alignment for Multi-Modal Prompt Learning}

\author{
    Jingsheng Gao\textsuperscript{\rm 1}, Jiacheng Ruan\textsuperscript{\rm 1},
    Suncheng Xiang\textsuperscript{\rm 1}, Zefang Yu\textsuperscript{\rm 1}\\
    Ke Ji\textsuperscript{\rm 2}, Mingye Xie\textsuperscript{\rm 1}, Ting Liu\textsuperscript{\rm 1}, Yuzhuo Fu\textsuperscript{\rm 1}\thanks{Corresponding Author.}
}
\affiliations{
    \textsuperscript{\rm 1} School of Electronic Information and Electrical Engineering, Shanghai Jiao Tong University, China \\
    \textsuperscript{\rm 2} School of Computer Science and Engineering, Southeast University, China \\
    \{gaojingsheng, jackchenruan, xiangsuncheng17, yuzefang, xiemingye, louisa\_liu, yzfu\}@sjtu.edu.cn \\
     keji@seu.edu.cn
}

\usepackage{bibentry}

\begin{document}

\maketitle

\begin{abstract}

    With the success of pre-trained visual-language (VL) models such as CLIP in visual representation tasks, transferring pre-trained models to downstream tasks has become a crucial paradigm. Recently, the prompt tuning paradigm, which draws inspiration from natural language processing (NLP), has made significant progress in VL field. However, preceding methods mainly focus on constructing prompt templates for text and visual inputs, neglecting the gap in class label representations between the VL models and downstream tasks. To address this challenge, we introduce an innovative label alignment method named \textbf{LAMM}, which can dynamically adjust the category embeddings of downstream datasets through end-to-end training. Moreover, to achieve a more appropriate label distribution, we propose a hierarchical loss, encompassing the alignment of the parameter space, feature space, and logits space. We conduct experiments on 11 downstream vision datasets and demonstrate that our method significantly improves the performance of existing multi-modal prompt learning models in few-shot scenarios, exhibiting an average accuracy improvement of 2.31(\%) compared to the state-of-the-art methods on 16 shots. Moreover, our methodology exhibits the preeminence in continual learning compared to other prompt tuning methods. Importantly, our method is synergistic with existing prompt tuning methods and can boost the performance on top of them. Our code and dataset will be publicly available at https://github.com/gaojingsheng/LAMM.
  
\end{abstract}

\section{Introduction}

\begin{figure}[htbp]
	\centering
	\includegraphics[width=0.44\textwidth, height=0.27\textwidth]{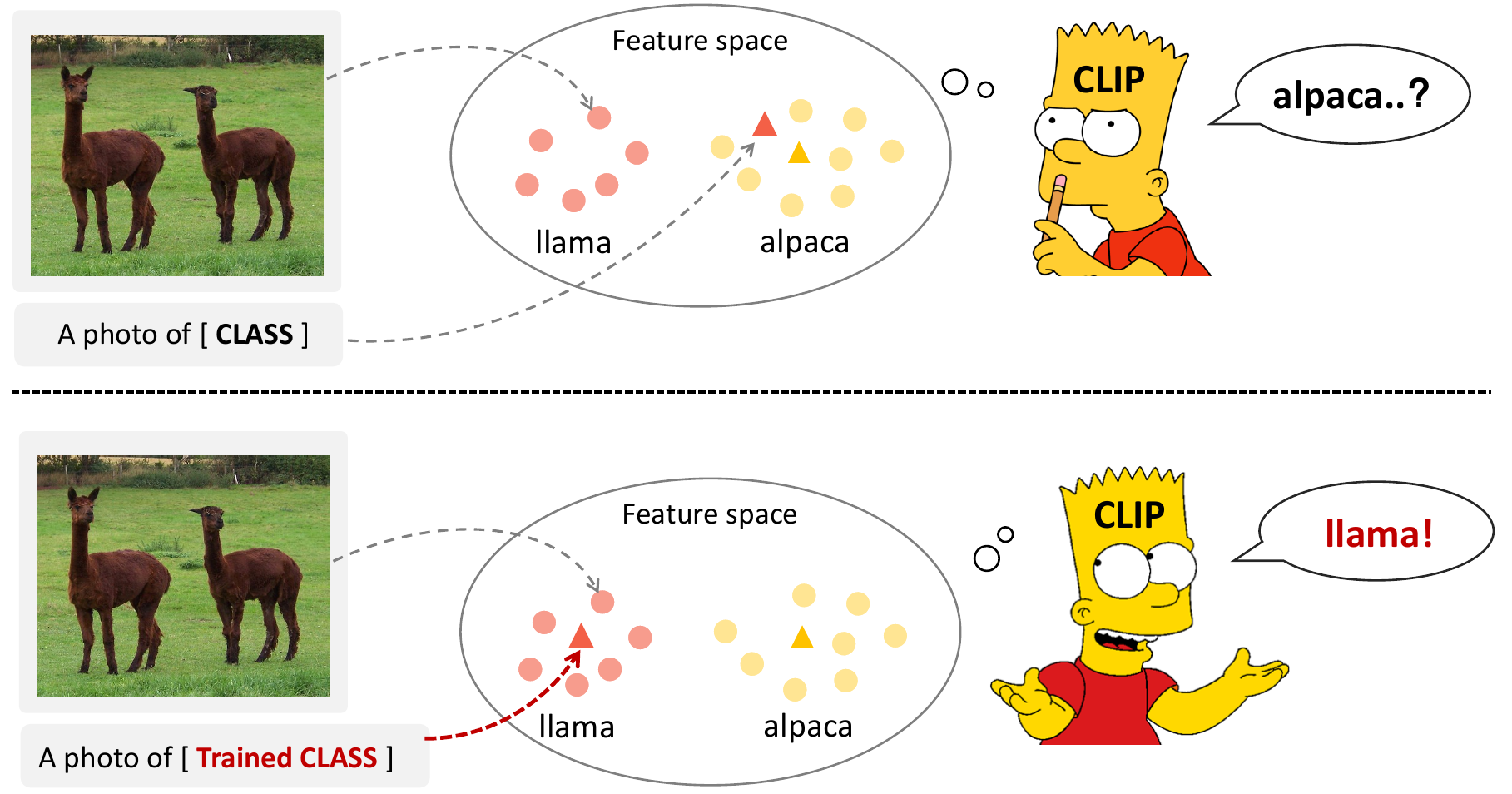}
	\caption{CLIP is more inclined to classify an image as belonging to a similar category. Altering the category text feature's position can enhance CLIP's recognition capabilities.}
	\label{fig:frontpic}
\end{figure}

Building machines to comprehend multi-modal information in real-world environments is one of the primary goals of artificial intelligence, where vision and language are the two crucial modalities~\cite{VLSurvey}. One effective implementation method is to pre-train a foundational vision-language (VL) model on a large-scale visual-text dataset and then transfer it to downstream application scenarios~\cite{CLIP, ALIGN}. Typically, VL models employ two separate encoders to encode image and text features, followed by the design of an appropriate loss function for training. However, finetuning on extensively trained models is costly and intricate, thus making the question of how to effectively transfer pre-trained VL models to downstream tasks a inspiring and valuable issue.




Prompt learning provides an effective solution to this problem, which provides downstream tasks with corresponding textual descriptions based on human prior knowledge and can effectively enhance the zero-shot and few-shot recognition capability of VL models. Through trainable templates with a small number of task-specific parameters, the process of constructing templates is further automated via gradient descent instead of manual constructions~\cite{PromptTuning}. Specifically, existing multi-modal prompt tuning methods~\cite{CoOp, CoCoOp, Maple} use the frozen CLIP~\cite{CLIP} model and design trainable prompts separately for the textual and visual encoders. These approaches ensure that VL models could be better transferred to downstream tasks without any changes to the VL model's parameters. However, their approach mainly focuses on the prompt template that is applicable to all categories, overlooking the feature representation of each category. 





The $<$CLASS$>$ token in the text template is crucial in classifying an image into the proper category. For example, as depicted in Figure \ref{fig:frontpic}, \textit{llamas} and \textit{alpacas} are two animals that resemble each other closely. In CLIP, there exists a propensity to misclassify a \textit{llama} as an \textit{alpaca} owing to the overrepresentation of \textit{alpaca} data in the pre-training dataset. By refining the text embedding position, CLIP can distinguish between these two species with trained feature space. Hence, identifying an optimal representation for each category in downstream tasks within the VL model is crucial. In the field of NLP, there exists the soft verbalizer~\cite{ProVerbalizer}, which enables the model to predict the representation of $<$MASK$>$ in the text template to represent the category of the original sentence on its own. Unlike NLP, it is infeasible to task the text encoder of the VL model with predicting the image category directly. Nevertheless, we can optimize the category embeddings of various categories within the downstream datasets to increase the similarity between each image and its corresponding category description. 




Consequently, we introduce a label alignment technique named LAMM, which automatically searches optimal $<$CLASS$>$ embeddings through gradient optimization. To the best of our knowledge, the concept of trainable category token is first proposed in the pre-trained VL models. Simultaneously, to prevent the semantic features of the entire prompt template from deviating too far, we introduce a hierarchical loss during our training phase. The hierarchical loss facilitates alignment of category representations among parameter, feature and logits spaces. With these operations, the generalization ability of CLIP model can be preserved in LAMM, which makes LAMM better distinguish different categories in downstream tasks while preserving the semantics of the original category descriptions. Furthermore, given that LAMM solely fine-tunes the label embeddings within the downstream dataset, it doesn't encounter the issue of catastrophic forgetting typically encountered in conventional methods during continual learning. 


 

We conduct experiments on 11 datasets, covering a range of downstream recognition scenarios. In terms of models, we test the vanilla CLIP, CoOp~\cite{CoOp}, and MaPLe~\cite{Maple}, which currently perform best in multi-modal prompt learning. Extensive experiments demonstrate the effectiveness of the proposed method within few-shot learning, illuminating its merits in both domain generalization and continual learning. Furthermore, our approach, being compatible to prevailing multi-modal prompt techniques, amplifies their efficacy across downstream datasets, ensuring consistent enhancement.

\section{Related Work}

\begin{figure*}[htbp]
	\centering
    \includegraphics[width=0.95\textwidth]{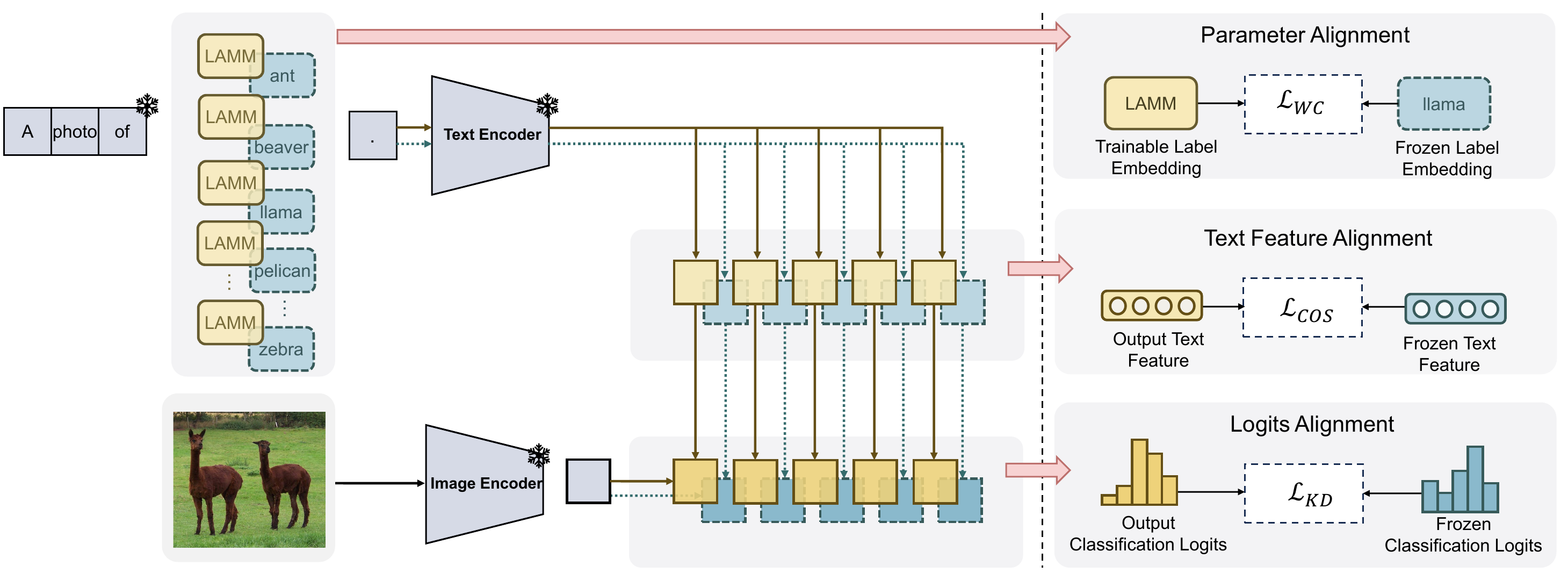}
	\caption{The whole architecture of LAMM. We replace the category tokens in the downstream dataset with trainable vectors and incorporate a hierarchical loss to preserve the CLIP's generalization ability of each category. The gray boxes represent the prompt template and frozen model, the blue boxes indicate the original label embeddings/features/logits, and the yellow ones denote the labe embeddings/features/logits during training.}
	\label{fig:CLIPLAMM}
\end{figure*}



\myparagraph{Vision Language Models} In recent years, the development of Vision-Language Pre-Trained Models (VL-PTMs) has made tremendous progress, as evidenced by models such as CLIP~\cite{CLIP}, ALIGN~\cite{ALIGN}, LiT~\cite{LIT} and FILIP~\cite{FILIP}. These VL-PTMs are pre-trained on large-scale image-text corpora and learn universal cross-modal representations, which are beneﬁcial for achieving strong performance in downstream VL tasks. For instance, CLIP are pre-trained on massive collections of image-caption pairs sourced from the internet, utilizing a contrastive loss that brings the representations of matching image-text pairs closer while pushing those of non-matching pairs further apart. After the pre-trained stage, CLIP has demonstrated exceptional performance on learning universal cross-modal representation in image-recognition~\cite{CLIP-adapter}, object detection~\cite{Detection2}, image segmentation~\cite{Segmentation1} and vision question answering~\cite{VQA1}.



\myparagraph{Prompt Learning} Prompt learning adapt the pre-trained language models (PLMs) by designing a prompt template to leverage the power of PLMs to unprecedented heights, especially in few-shot settings~\cite{liu2021pre}. With the emergence of large-scale VL models, numerous researchers have attempted to integrate prompt learning into VL scenarios, resulting in the development of prompt paradigms that are better adapted for these scenarios. CoOp~\cite{CoOp} first introduces the prompt tuning approach to VL models by learnable prompt template words. Co-CoOp~\cite{CoCoOp} improves the performance on novel classes by incorporating instance-level image information. Unlike prompt engineering in text templates, VPT~\cite{VPT} inserts learnable parameters into the vision encoder. MaPLe~\cite{Maple} appends a soft prompt to the hidden representations at each layer of the text and image encoders, resulting in a new solid performance in few-shot image recognition.

Previous multi-modal prompts have primarily focused on the engineering of prompt templates for textual and visual inputs while neglecting the significance of label representations in such templates. However, label verbalizers have already been proven to be effective in few-shot text classification~\cite{ManualVerbalizer}, where verbalizers aim to reduce the gap between model outputs and label words. To alleviate the required expertise and workload for constructing a manual verbalizer, Gao et al.~\cite{Gaoverb} design search-based methods for better verbalizer choices during the training optimization process. Some other researches~\cite{SoftVerbalizer, ProVerbalizer} propose trainable vectors as soft verbalizers to replace label words, which eliminates the difficulty of searching the entire dictionary. 


Hence, we introduce trainable vectors to substitute label words in multi-modal prompts. Our approach aims to align label representations in the downstream datasets with pre-trained VL models, reducing the discrepancy in category descriptions between downstream datasets and VL-PTMs. 






\section{Methodology}

In this section, we will introduce how our proposed LAMM can be incorporated into CLIP seamlessly, accompanied by our hierarchical loss. The whole architecture of LAMM and hierarchical alignment is shown in Figure \ref{fig:CLIPLAMM}.


\subsection{Preliminaries of CLIP}
\label{preliminaries}

 
CLIP is a VL-PTM comprising a vision encoder $\phi$ and a text encoder $\psi$. These two encoders extract image and text information, respectively, and map them to a common feature space $R^d$, where the two feature spaces align well.

Given an input image $x$, the image encoder will extract the corresponding image representation ${I}_{x} = \phi(x)$. For each downstream dataset, there will be ${k}$ classes and each class will be filled in a manual prompt template, such as “a photo of $<$CLASS$>$”. Then the text encoder will generate feature representations for each category by further processing, resulting in each category feature. During training, CLIP maximizes the cosine similarity between image representation and its corresponding category representation, while minimizing the cosine similarity between unmatched pairs. During zero-shot inference, the prediction probability of $i$-th category is computed as:
\begin{equation}
\label{eq:yi}
p(y=i \mid I)=\frac{\exp \left(\cos \left(I_{x},\psi(y_i)\right) / \tau\right)}{\sum_{j=1}^{k} \exp \left(\cos \left(I_{x}, \psi(y_j)\right) / \tau\right)}
\end{equation}
where $\tau$ is a temperature parameter acquired by CLIP, while function $\cos$ represents cosine similarity.

\subsection{Label Alignment}
\label{sec: alignment}

Although CLIP has strong zero-shot performance, providing downstream tasks with corresponding textual descriptions can effectively enhance the zero-shot and few-shot recognition capability of CLIP. Previous prompt tuning work on text template mainly focus on the training of “a photo of”, while neglecting the optimization of “$<$CLASS$>$”. To effectively align class labels in downstream tasks to pre-trained models, we propose LAMM, which automatically optimizes the label embedding through end-to-end training. We take CLIP as an example, LAMM on CLIP only finetunes the class embedding representation for downstream tasks. In this way, the prompt template of LAMM converts to:
\begin{equation}
    \label{eq:zi}
    z_i = [a] [photo] [of] [<M_i>] [.]
\end{equation}
where $<M_i>$ (i=1, 2, ..., k) represents a learnable token of the $i$-th category. Similar to Equation \ref{eq:yi}, the prediction probability of LAMM is computed as:
\begin{equation}
p(y=i \mid I)=\frac{\exp \left(\cos \left(I_{x},\psi(z_{i})\right) / \tau\right)}{\sum_{j=1}^{k} \exp \left(\cos \left(I_{x}, \psi(z_{i})\right) / \tau\right)}
\end{equation}

During training, we only update the category vectors $\{<M_i>\}_{i=1}^{k}$ in each downstream dataset, which will decrease the gap between image representation and its corresponding category representation.

Futhermore, LAMM can be applied into existing multi-modal prompting methods. Take CoOp for example, the difference between CoOp and vanilla CLIP is replacing the prompt template with M learnable tokens. Thus, the prompt template of CoOp+LAMM becomes:
\begin{equation}
    z_i^{*} = [V_1] [V_2] ... [V_4] [<M_i>]
\end{equation}

By replacing the class representation with a trainable vector, we can integrate our method into any existing multi-modal prompt approach. 

\subsection{Hierarchical Loss}

A well-aligned feature space is the key of strong zero-shot ability of CLIP, which also facilitates the learning of downstream tasks. Given that LAMM does not introduce any modifications or additions to the CLIP model's parameters, the image representation within the aligned feature space remains fixed, while the trainable embedding of each category changes the text representation within the feature space. Nevertheless, a single text representation corresponds to multiple images of a given category in the downstream datasets, despite the entire training process being conducted under few-shot settings. This situation may lead to overfitting of the trainable class embeddings to the limited number of images in the training set. Hence, we propose a hierarchical loss (HL) to safeguard the generalization ability in the parameter space, feature space and logits space. The rationale for incorporating the loss function is to ensure the alignment of our model with the highly adaptable zero-shot CLIP model, which demonstrates robust generalization capabilities across diverse dimensions.

\myparagraph{Parameter Space} To mitigate the risk of overfitting in models, numerous machine learning techniques employ parameter regularization to enhance the generalization ability on unseen samples. In this regard, the weight consolidation (WC) ~\cite{WCLoss} loss is employed as follows:
\begin{equation}
\mathcal{L}_{\mathrm{WC}}=\sum_i\left(\theta_i-\bar{\theta}_i\right)^2 
\end{equation}
where $\theta$ is the trainable parameters of the current model, and $\bar{\theta}$ is the reference ones. In LAMM, $\theta$ represents the trainable label embeddings, whereas $\bar{\theta}$ represents the original label embeddings. While parameter regularization can address the issue of overfitting, excessive regularization may hinder the model's ability to adequately capture the features and patterns present in the training data. We set the coefficient of $\mathcal{L}_{\mathrm{WC}}$ inversely proportional to the number of training shots, suggesting that fewer shots will be accompanied by a more strict regularization process within parameter space.

\myparagraph{Feature Space} In addition to the parameter space, it is crucial for the text feature of our trained category to align with the characteristics of the training images. During the training process, the text feature of our trained category gradually converges towards the characteristics present in the training images. However, if the representations of the few-shot training images for a particular category do not align with those of the entire image dataset, it can lead to the label semantics overfitting to specific samples. For example, consider the label embedding of the "llama" image illustrated in Figure \ref{fig:CLIPLAMM}. The label embedding may overfit to the background grass information, even though llamas may not always be associated with grass in all instances. In order to mitigate the overfitting of text features for each category, we employ a text feature alignment loss to restrict the optimization region of the text feature. Drawing inspiration from previous work on similarity at the semantic level~\cite{SimCSE}, we employ a cosine similarity loss for alignment. In this approach, for a category template $z_i$ in Equation \ref{eq:zi}, the original prompt template $y_i$ from Equation \ref{eq:yi} serves as the center of its optimization region. The cosine loss is formulated as follows:
\begin{equation}
    \mathcal{L}_{COS}= \sum_i 1- \cos (\psi(z_i), \psi(y_i))
\end{equation}

\myparagraph{Logits Space} The strong generalization ability of CLIP plays a crucial role in the effectiveness of multi-modal prompting methods within few-shot scenarios. While the previous two losses enhance the generalization capability of LAMM through regularization in the parameter and feature spaces, it is desirable to minimize the distribution shift of logits between image representations and different text representations, as compared to the zero-shot CLIP. Therefore, we introduce a knowledge distillation loss in the classification logits space, which allows for the transfer of generalization knowledge from CLIP to LAMM. The distillation loss can be formulated as follows:
\begin{equation}
\mathcal{L}_{\mathrm{KD}}=-\sum \cos \left(I_x, \psi\left(z_i\right)\right)  \log \left(\cos \left(I_x, \psi\left(y_i\right)\right)\right)
\end{equation}
\myparagraph{Total Loss} To train the LAMM for downstream tasks, cross-entropy (CE) loss is applied to the similarity score as same to finetuning the CLIP model:
\begin{equation}
    \mathcal{L}_{\mathrm{CE}}=\frac{1}{N} \sum_{i=1}^N \mathrm{CE}\left(\tau \cdot \cos (I_{x}, \psi(z_{i})), y_i\right)
\end{equation}
where $\tau$ is a parameter learned during the pre-training. In this way, the total loss is:
\begin{equation}
\mathcal{L}=\mathcal{L}_{C E}+\lambda_1 * \mathcal{L}_{W C}+\lambda_2 * \mathcal{L}_{C O S}+\lambda_3 * \mathcal{L}_{\mathrm{KD}}
\end{equation}

where $\lambda_1, \lambda_2, \lambda_3$ are hyper-parameters. To prevent the redundancy of adjusting parameter, we set $\lambda_1=1/n, \lambda_2=1, \lambda_3=0.05$ for all our experiments empirically, where $n$ represents the number of training shots.

\section{Experiments}

\begin{figure*}[ht]
	\centering
        \includegraphics[width=0.92\textwidth]{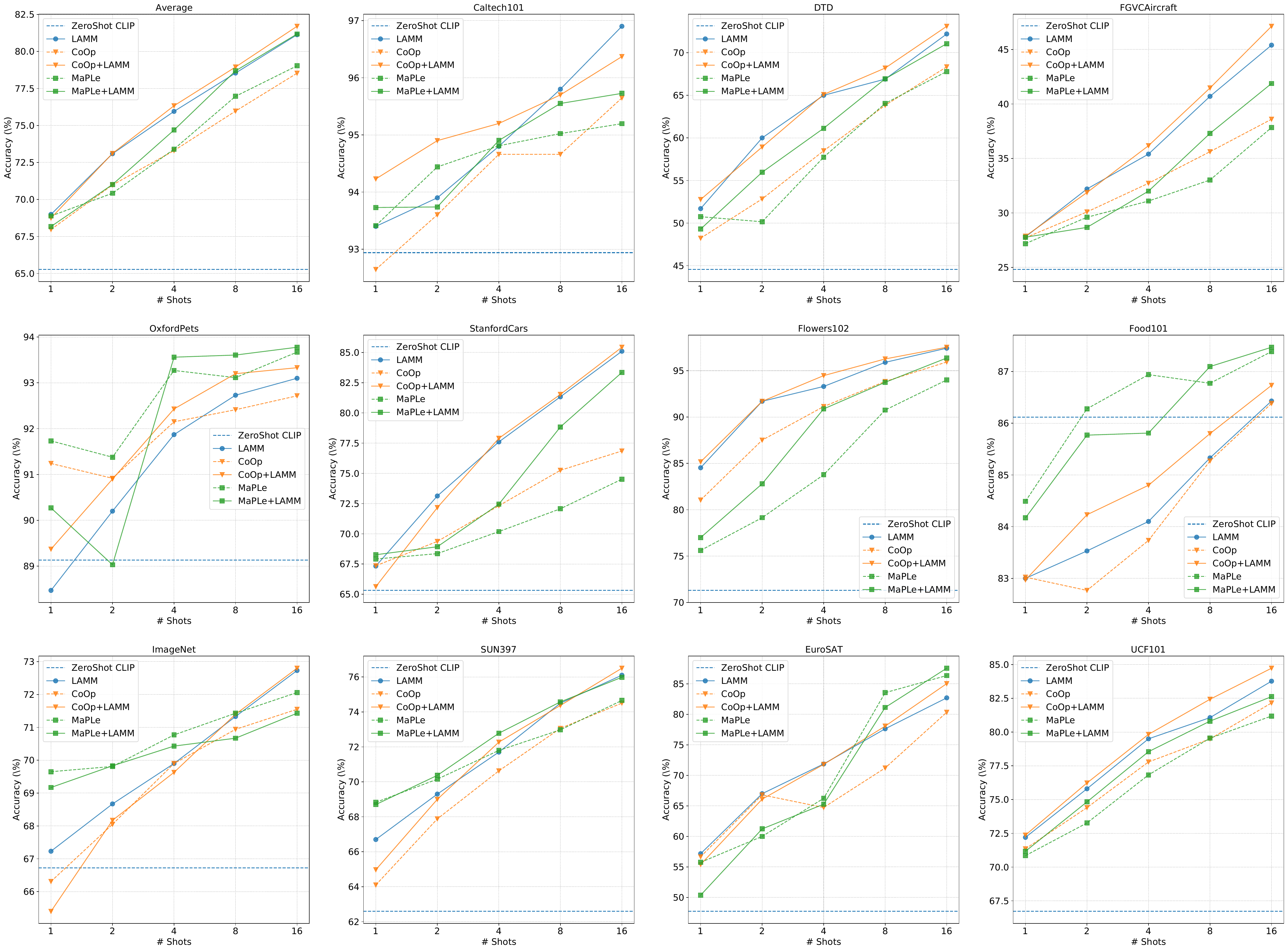}
	\caption{Main results over 11 datasets under the few-shot learning setting. We report the average accuracy (\%) of 1/2/4/8/16 shots over three runs. Overall, the proposed LAMM enhances the performance of CLIP, CoOp, and MaPLe.}
	\label{fig:MainResults}
\end{figure*}

    

\subsection{Few-shot Settings}

\myparagraph{Datasets} We follow the datasets used in previous works~\cite{CoOp, Maple} and evaluate our method on 11 image classification datasets, including Caltech101~\cite{Caltech101}, ImageNet~\cite{ImageNet}, OxfordPets~\cite{Oxford_pets}, StanfordCars~\cite{Stanford_cars}, Flowers102~\cite{Oxford_flowers}, Food101~\cite{Food101}, FGVCAircraft~\cite{FGVC_aircraft}, SUN397~\cite{SUN397}, UCF101~\cite{UCF101}, DTD~\cite{DTD} and EuroSAT~\cite{Eurosat}. Besides, we follow ~\cite{CLIP, CoOp} to set up the few-shot evaluation protocol for our few-shot learning experiments. Specifically, we use 1, 2, 4, 8, and 16 shots for training respectively, and evaluate the models on the full test sets. All experimental results are the average of the results obtained from running the experiments on seeds 1, 2 and 3.


\myparagraph{Baselines} We compare the results across LAMM, CoOp, and MaPLe. Furthermore, we incorporate LAMM into CoOp and MaPLe to prove the compatibility of LAMM. To maintain fair controlled experiments, all prompt templates in our experiments are initialized from "a photo of $<$CLASS$>$". Besides, the pre-trained model adopted here is ViT-B/16 CLIP since MaPLe can only be adopted in transformer-based VL-PTMs. We keep the same training parameters (e.g., learning rate, epochs, and other prompt parameters) of each model in their original settings, where the epoch of CoOp is 50 and MaPLe is 5. As for vanilla CLIP + LAMM, we follow the settings of CoOp. All of our experiments are conducted on a single NVIDIA A100. The corresponding hyper-parameters are ﬁxed across all datasets in our work. Moreover, adjusting parameters for different training shots and datasets can boost performance.

\subsection{Comparision to the State-of-the-art Methods}

\myparagraph{Main Results on 11 Datasets.} We compare LAMM, CoOp, MaPLe and zero-shot CLIP on the 11 datasets as mentioned above, demonstrated in in Figure \ref{fig:MainResults}.  We can observe that LAMM yield best performance among all shots compared to the state-of-the-art multi-modal prompt methods. LAMM only replaces $<$CLASS$>$ for each category with a trainable vector, while CoOp replaces "a photo of" with trainable vectors. However, compared to CLIP, LAMM demonstrates an enhancement of +1.02, +2.11, +2.65, +2.57, and +2.57(\%), on the 1, 2, 4, 8, and 16 shots, respectively. The preceding result highlights the importance of finetuning label embeddings for downstream tasks, compared to the previous focus solely on prompt template learning in multi-modal prompt learning. It suggests that label embeddings are even more crucial than prompt templates for pre-trained models' transferability to downstream tasks. Furthermore, as the number of shot increases, the improvement observed with LAMM becomes more clear. This can be attributed to the challenge LAMM faces in learning representative embeddings for categories in downstream tasks when provided with fewer shots.

\begin{table}[ht]
    \centering
    \footnotesize
    \resizebox{0.46\textwidth}{!}{%
    \begin{tabular}{l|c|cc|c}
    \toprule
    \multirow{2}{*}{Method} & Source & \multicolumn{2}{c|}{Target} & \multirow{2}{*}{Average} \\

     & ImageNet & -V2 & -Sketch &\\
    \midrule

    CLIP & 66.73 & 60.83 & 46.15 & 57.90 \\

    CoOp & 71.51 & 64.20 & 47.99 & 61.23 \\

    CoCoOp & 71.02 & 64.07 & 48.75 & 61.28 \\

    
    MaPLe & 70.02 & 64.07 & \textbf{49.15} & 61.08 \\

    LAMM & \textbf{72.73}  & \textbf{65.13}  & 48.11 & \textbf{61.99} \\

    \bottomrule
    \end{tabular}
    }
    \caption{Comparison of LAMM with existing methods in domain generalization setting. LAMM shows highest performance on average.}
    \label{tab:robustness} 
\end{table}



\myparagraph{Domain Generalization} We evaluate the cross-dataset generalization ability of LAMM by training it on ImageNet and evaluating on ImageNetV2~\cite{imagenetv2} and Imagenet-Sketch~\cite{imagenet-sketch}, following~\cite{Maple}. The evaluating datasets have same categories with the training set. But the three datasets are different in domain distribution. The experimental results are shown in Table \ref{tab:robustness}. LAMM achieves the best performance, which surpasses MaPLe [51] 1.06\% on ImageNet-V2 and falls behind MaPLe 1.04\% on ImageNet-Sketch. On average, LAMM exceeds MaPLe 0.91\%. This indicates the LAMM is also capable of out-of-distribution tasks.

\subsection{Combination with State-of-the-art methods}
\begin{table*}
    \centering
    \footnotesize
    \resizebox{0.93\textwidth}{!}{%
    \begin{tabular}{cc|cccccccccccccc}
    \toprule
     \rotatebox{60}{Method} & \rotatebox{60}{Subset} &\rotatebox{60}{ImageNet} &  \rotatebox{60}{Caltech101}  & \rotatebox{60}{DTD} & \rotatebox{60}{FGVCAircraft}  &  \rotatebox{60}{StandfordCars}  &  \rotatebox{60}{Flowers102}   &  \rotatebox{60}{OxfordPets} &  \rotatebox{60}{Food101}  & \rotatebox{60}{EuroSAT} & \rotatebox{60}{UCF101} & \rotatebox{60}{SUN397} & \rotatebox{60}{Average} &  \rotatebox{60}{Degradation} \\ 

    
    \midrule
    
    \multirow{2}{*}{Zero-shot CLIP} & Set1 & 72.43 & 96.84 & 53.24 & 27.19 & 63.37 & 72.08 & 91.17 & 90.10 & 56.48 & 70.53 & 69.36 & 69.34 & 69.34 (\textbf{0}) \\

    & Set2 & 68.14 & 94.00 & 59.90 & 36.29 & 74.89 & 77.80 & 97.26 & 91.22 & 64.05 & 77.50 & 75.35 & 74.22 & \\

    \midrule
    
    \multirow{2}{*}{CoOp} & Set1 & 71.77 & 97.07 & 54.63 & 23.40 &  61.63 & 61.97 & 93.63 & 87.10 & 72.37 & 71.03 & 72.53 & 69.74 & 82.69 (\textbf{-12.95}) \\

    & Set2 &  73.67 &   96.37 & 76.57& 54.53 &  87.07 & 97.50 & 97.70 &  91.47 &  90.27 & 88.13 & 83.77 & 85.18 & \\

    \midrule
    
    \multirow{2}{*}{MaPLe} & Set1 & 74.37 & 97.10 & 70.83 & 30.27 & 65.77 & 77.53 & 95.03 & \textbf{90.6} & 82.73 & 78.17 &  77.60 & 76.36 & 82.28 (\textbf{-5.92})\\

    & Set2 & 74.13 & \textbf{96.50} & 77.40 & 53.73 & 83.27 & 97.00 & \textbf{98.17}  & \textbf{92.27} & \textbf{92.60} & 88.33 &  83.95 &  85.21 &  \\

    \midrule
    
    \multirow{2}{*}{LAMM} & Set1 &  \textbf{77.23} & \textbf{98.40} & \textbf{83.30} & \textbf{43.27} & \textbf{81.63} & \textbf{97.80} &  \textbf{95.17} & 89.83 &  \textbf{90.60} &  \textbf{86.17} & \textbf{82.13} & \textbf{84.14} & 84.14 (\textbf{0})  \\

    & Set2 & \textbf{74.57} & 96.03 & \textbf{79.47} & \textbf{61.47} & \textbf{91.97} & \textbf{98.33} & 97.93 & 91.73 & 91.80 & \textbf{89.23} & \textbf{84.80} & \textbf{87.03} & \\

    \bottomrule
    \end{tabular}
    }
    \caption{Comparison of LAMM and other prompting methods on 16-shot classs incremental learning. Initially, models are trained on Set 1, followed by training on Set 2. The contents within the parentheses of term "Degradation" refers to the decline in evaluation results on Set 1 subsequent to further training on Set 2.}
    \label{tab:incremental} 
    
\end{table*}

\myparagraph{The improvement of incorporating LAMM} We compare the performance of the existing methods CoOp and MaPLe before and after integrating our approach, LAMM, which uses trainable vectors and a hierarchical  loss function. From Figure \ref{fig:MainResults}, we observe that LAMM significantly improves the performance of existing models in few-show scenarios, except for 1-shot learning. For CoOp, the average accuracy variations across 11 datasets with 1, 2, 4, 8, and 16 shots are +0.77, +2.13, +3.03, +2.98, +3.17(\%), respectively. As for MaPLe, the variations are -0.71, +1.79, +1.29, +1.72, +2.15(\%). The incorporation of LAMM has yielded exceedingly favorable outcomes across all three methodologies, reaffirming LAMM's status as a reliable and exceptional few-shot learner.

Besides, the enhanced performance of LAMM improves significantly with the increase in shots, and the reason behind this is that our approach primarily operates on label representation. When the number of training sample is small, such as in the case of 1-shot learning, the label representation learns numerous specific features associated with the single image, which constitutes a significant portion of noise rather than the representation of the entire category. For instance, the significant reduction in accuracy of MaPLe with LAMM in the case of 1-shot learning is primarily due to the decrease of 5.38\% accuracy on the EuroSAT~\cite{Stanford_cars}. This is owing to the fact that the EuroSAT is a satellite-image dataset, which suggests that the gap between the feature spaces of images and text is considerable. Consequently, it makes LAMM more susceptible to overfitting on the features of limited-sample images. Moreover, MaPLe has more trainable parameters than CoOp, which makes it more prone to overfitting than CoOp.


\myparagraph{Comparisons of different LAMMs} After incorporating LAMM into previous methods, we evaluate the performance of LAMM, CoOp+LAMM, and MaPLe+LAMM. From Table \ref{tab:loss}, we can observe that CoOp+LAMM achieving the best results. To be more specific, LAMM achieves a reduction of +0.25, -0.02, -0.38, -0.41, -0.60(\%) compared to CoOp+LAMM on 1, 2, 4, 8, and 16 shots. We conclude that CoOp+LAMM can achieve superior results due to CoOp's ability to search a soft template that is more effective than "a photo of," without significantly altering the semantic premise of the prompt template. In contrast, MaPLe introduces more trainable parameters, which may result in a greater deviation of the semantic meaning of the soft prompt template from that of "a photo of". 


\subsection{Class Incremental Learning}

Since LAMM exclusively modifies the training class embedding without altering any parameters of the CLIP model, it is confined to using the unmodified CLIP and the untrained novel class embedding during base-to-novel testing. In this way, LAMM's performance for novel classes is same to that of the zero-shot CLIP. However, except MaPLe exceed zero-shot CLIP 0.92\% on novel class testing, other prompt tuning methods falls behind zero-shot CLIP.

Apart from base-to-novel testing, incremental learning is also an significant issue when aiming to broaden a model's knowledge base. Following MaPLe, we partition the datasets into base and novel classes, designating the base classes as Set 1 and the novel ones as Set 2. Initially, each model is trained on Set 1, after which the continual training persists on Set 2. Finally, we evaluate the performance of each model on both Set 1 and Set 2. The results are shown in Table \ref{tab:incremental}. The term 'Degradation' the disparity in performance on Set 1 prior to and subsequent to incremental training on Set 2. Such a decline represents the forgetting of prior  tasks during the incremental learning process. 

It is evident that LAMM exhibits superior performance on both Set 1 and Set 2, particularly on Set 1. The observed decline in performance of CoOp and MaPLe on Set 1 can be attributed to the phenomenon of forgetting during continual learning. Especially for CoOp, its performance on Set 1 closely approximates that of zero-shot CLIP. This indicates that the adaptable text template is highly sensitive to variations in testing tasks. Consequently, CoOp necessitates an entirely new template upon encountering novel classes. Although MaPLe outperforms CoOp, it still undergoes an average degradation of -5.92\%.


As LAMM solely manipulates the embedding of new classes while preserving the existing class embeddings, the performance of LAMM on previous classes remains stable during continual training on new classes. Conversely, in the case of CoOp and MaPLe, when the number of categories to be learned increases, the model's performance will significantly decline if retraining from scratch on all categories is not undertaken. This unique prowess of LAMM in incremental learning implies its suitability and desirability for deployment in downstream applications.

\subsection{Ablation Experiments}

\myparagraph{Influence of Hierarchical Loss} To validate the effect of the proposed loss function, we conduct a comprehensive study on three models among all the 11 datasets. For comparison, we consider LAMM, CoOp+LAMM, and MaPLe+LAMM with or without hierarchical loss. Table \ref{tab:loss} presents the averaged results over three runs. We observe that the results exhibit a significant difference upon the introduction of the loss, with hierarchical loss proving to be highly essential in enhancing the performance of LAMM. Furthermore, the degree of improvement brought by hierarchical loss becomes more apparent when there are fewer shots involved. This is due to that finetuning to align the label with few samples, especially with a single image, can result in the label overfitting to other noise information within the image. Since hierarchical loss incorporates strong constraints over parameter space, text feature space and logits space, which ensures that the semantic meaning of the label does not deviate too far from its original semantic meaning after training, it can improve LAMM's performance more significantly with fewer samples compared to without hierarchical loss.

\begin{table}[ht]

    \centering
    \footnotesize
    \resizebox{0.47\textwidth}{!}{%
    \begin{tabular}{l|ccccc}
    \toprule
    Method & 1-shot & 2-shot & 4-shot & 8-shot & 16-shot \\
    
    \midrule
    $\textbf{Zero-shot CLIP} $ & 65.27 & 65.27 & 65.27 & 65.27 & 65.27  \\
    $\textbf{LAMM}{\textbf{\textit{(w/o HL)}}} $ & 61.39 & 66.8 & 72.13 & 76.21 & 79.93 \\
    $\textbf{LAMM} $ & \textbf{68.99} & 73.09 &	75.95 &	78.54 &	81.13\\
    \midrule
    $\textbf{CoOp}$ & 67.97 & 70.98 & 73.3 & 75.97 & 78.53 \\
    $\textbf{CoOp+LAMM}{\textbf{\textit{(w/o HL)}}} $ & 61.23 & 66.33 & 72.06 & 76.19 & 79.94 \\
    $\textbf{CoOp+LAMM} $ & 68.74 & \textbf{73.11}	& \textbf{76.33} & \textbf{78.95}	& \textbf{81.71} \\
    \midrule
    $\textbf{MaPLe} $ & 68.88 & 69.22 & 73.4 & 76.97 & 79.03 \\
    $\textbf{MaPLe+LAMM}{\textbf{\textit{(w/o HL)}}} $ & 61.25 & 67.70 & 73.39 & 77.64 & 80.74 \\
    $\textbf{MaPLe+LAMM} $  & 68.17 & 	71.01 & 74.69 & 78.69 & 81.18 \\
    \bottomrule
    \end{tabular}
    }
    \caption{ Comparision of with or without hierarchical loss among vanilla CLIP, CoOp, and MaPLe. }
    \label{tab:loss}

\end{table}

\myparagraph{Ablations of Hierarchical Loss} In the context of the hierarchical loss, it is composed of losses from three dimensions: parameter space, feature space, and logits space. Therefore, we conducted an ablation study on these varying levels of loss in the context of 16-shot, as illustrated in Table \ref{hierloss}. It is evident that each of the three losses contributes positively to the final outcome. However, their cumulative effects cannot be simply combined, given that the loss from one feature space can influence the characteristics of another space. The universal objective of these losses is to retain the generalization capability of CLIP within LAMM, preventing overfitting on specific samples.

\begin{table}[htbp]
    \begin{minipage}[c]{0.48\linewidth}
        \centering
        \footnotesize
        \setlength{\tabcolsep}{2pt}
        \resizebox{0.95\textwidth}{!}{%
        \begin{tabular}{cccc|c}
        \hline
        $\mathcal{L}_{CE}$ & $\mathcal{L}_{WC}$ & $\mathcal{L}_{COS}$ & $\mathcal{L}_{KD}$ & Average\\ \hline
        \checkmark & & & & 79.93
         \\
        \checkmark & \checkmark & \checkmark & \checkmark & 81.13 \\ \hline
        \checkmark & & \checkmark & \checkmark & 80.98 \\
        \checkmark & \checkmark & & \checkmark &  80.86 \\
        \checkmark & \checkmark & \checkmark & & 81.08\\ \hline
        \checkmark & & & \checkmark & 80.82 \\
        \checkmark & & \checkmark & & 80.75 \\
        \checkmark & \checkmark & & & 80.63 \\ \hline
        \end{tabular}
        }
        \caption{Ablations on hierarchical loss function on 16-shot.}
        \label{hierloss}
    \end{minipage}
        \hfill
        \begin{minipage}[c]{0.48\linewidth}
        \centering
        \footnotesize
        \setlength{\tabcolsep}{2pt}
        \centering
        
        \begin{tabular}{c|cc}
        \toprule
        Setting &   Random  & Words  \\
        
        \midrule
        1-shot & 62.71 & 68.99 \\
        2-shot & 62.71 & 73.09  \\
        
        4-shot &  71.68 &  75.95\\
        
        8-shot & 76.65 & 78.54\\
        16-shot & 80.71 & 81.13 \\
        
        \bottomrule
        \end{tabular}
        \caption{Comparision of initializations from random and category words.}
        \label{tab:Initialization} 
    
    \end{minipage}

\end{table}

\myparagraph{Initialization} We conduct a comparison between initialization from category words and random initialization. The former uses the original word embedding of each category as the initialized category embedding, while the latter randomly initializes the category embeddings. Table \ref{tab:Initialization} shows the results, which are conducted on LAMM among 11 datasets and suggest that random initialization is inferior to category word initialization. This illustrates that training representations for each category from scratch in few-shot training is quite challenging, and this is reflected in the increasing discrepancy between random initialization and word initialization as the shot number decreases.


\begin{figure}[htbp]
	\centering
	\includegraphics[width=0.39\textwidth, height=0.28\textwidth]{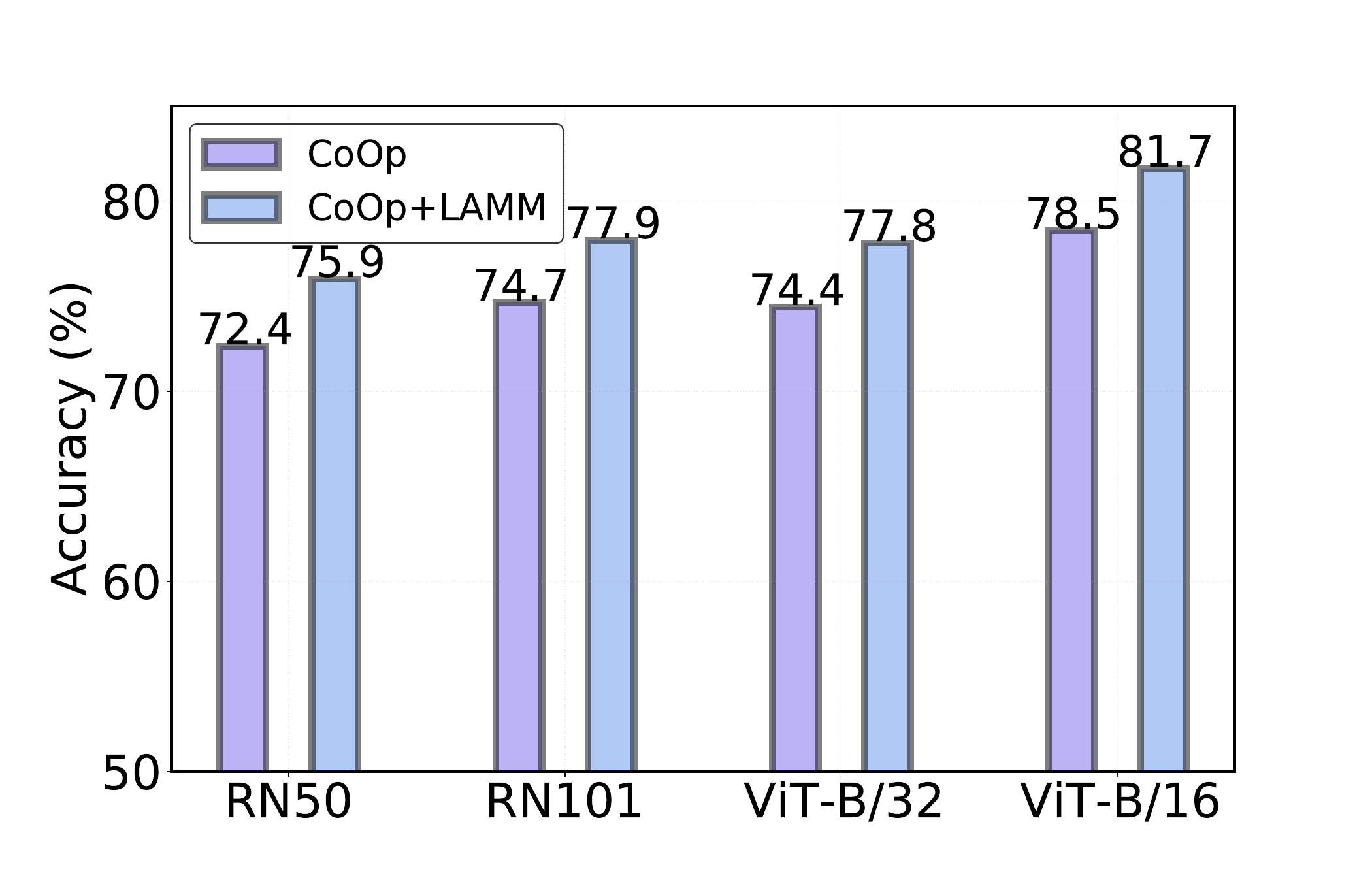}
	\caption{Ablations among different vision backbones}
	\label{fig:backbones}
\end{figure}


\myparagraph{Vision Backbone} Figure \ref{fig:backbones} illustrates the average results on 11 datasets of various visual backbones, including CNN and ViT architectures. The results demonstrate that LAMM can consistently improve the performance of CoOp across different visual backbones. This further indicates that LAMM can enhance the VL model's transferability to downstream tasks in a stable manner.


\section{Conclusion}

Compared to traditional few-shot learning methods, prompt learning based on VL-PTMs has demonstrated strong transferability in downstream tasks. Our research reveals that in addition to prompt template learning, reducing the gap between VL-PTMs and downstream task label representations is also a significant research issue. Our paper provides a comprehensive study on how to align label representations in downstream tasks to VL-PTMs in a plug-and-play way. Our proposed LAMM has demonstrated a significant improvement in the performance of previous multi-modal prompt methods in few-shot scenarios. In particular, by simply incorporating LAMM into vanilla CLIP, we can achieve better results than previous multi-modal prompt methods. LAMM also demonstrates robustness in out-of-distribution scenarios, along with its superiority in the incremental learning process. These findings further demonstrate the immense potential of optimizing the transfer of VL-PTMs to downstream tasks, not only limited to image recognition, but also encompassing visually semantic tasks such as image segmentation, object detection, and more. We hope that the insights gained from our work on label representation learning will facilitate the development of more effective transfer methods for VL-PTMs.




\section*{Acknowledgments}

This research was supported by the National Natural Science Foundation of China under Project (Grant No.61977045).

\bibliography{aaai24}

\begin{thebibliography}{34}
\providecommand{\natexlab}[1]{#1}

\bibitem[{Bossard, Guillaumin, and Gool(2014)}]{Food101}
Bossard, L.; Guillaumin, M.; and Gool, L.~V. 2014.
\newblock Food-101 - Mining Discriminative Components with Random Forests.
\newblock In \emph{Proc. of ECCV}.

\bibitem[{Cimpoi et~al.(2014)Cimpoi, Maji, Kokkinos, Mohamed, and
  Vedaldi}]{DTD}
Cimpoi, M.; Maji, S.; Kokkinos, I.; Mohamed, S.; and Vedaldi, A. 2014.
\newblock Describing Textures in the Wild.
\newblock In \emph{Proc. of CVPR}.

\bibitem[{Cui et~al.(2022)Cui, Hu, Ding, Huang, and Liu}]{ProVerbalizer}
Cui, G.; Hu, S.; Ding, N.; Huang, L.; and Liu, Z. 2022.
\newblock Prototypical Verbalizer for Prompt-based Few-shot Tuning.
\newblock In \emph{Proc. of ACL}.

\bibitem[{Deng et~al.(2009)Deng, Dong, Socher, Li, Li, and Fei-Fei}]{ImageNet}
Deng, J.; Dong, W.; Socher, R.; Li, L.-J.; Li, K.; and Fei-Fei, L. 2009.
\newblock Imagenet: A large-scale hierarchical image database.
\newblock In \emph{Proc. of CVPR}.

\bibitem[{Du et~al.(2022)Du, Liu, Li, and Zhao}]{VLSurvey}
Du, Y.; Liu, Z.; Li, J.; and Zhao, W.~X. 2022.
\newblock A Survey of Vision-Language Pre-Trained Models.
\newblock In \emph{Proc. of IJCAI}.

\bibitem[{Fei{-}Fei, Fergus, and Perona(2007)}]{Caltech101}
Fei{-}Fei, L.; Fergus, R.; and Perona, P. 2007.
\newblock Learning generative visual models from few training examples: An
  incremental Bayesian approach tested on 101 object categories.
\newblock \emph{Comput. Vis. Image Underst.}

\bibitem[{Gao et~al.(2021)Gao, Geng, Zhang, Ma, Fang, Zhang, Li, and
  Qiao}]{CLIP-adapter}
Gao, P.; Geng, S.; Zhang, R.; Ma, T.; Fang, R.; Zhang, Y.; Li, H.; and Qiao, Y.
  2021.
\newblock CLIP-Adapter: Better Vision-Language Models with Feature Adapters.
\newblock \emph{CoRR}.

\bibitem[{Gao, Fisch, and Chen(2021)}]{Gaoverb}
Gao, T.; Fisch, A.; and Chen, D. 2021.
\newblock Making Pre-trained Language Models Better Few-shot Learners.
\newblock In \emph{Proc. of ACL}.

\bibitem[{Gao, Yao, and Chen(2021)}]{SimCSE}
Gao, T.; Yao, X.; and Chen, D. 2021.
\newblock Simcse: Simple contrastive learning of sentence embeddings.
\newblock \emph{arXiv preprint arXiv:2104.08821}.

\bibitem[{Hambardzumyan, Khachatrian, and May(2021)}]{SoftVerbalizer}
Hambardzumyan, K.; Khachatrian, H.; and May, J. 2021.
\newblock {WARP:} Word-level Adversarial ReProgramming.
\newblock In \emph{Proc. of ACL}.

\bibitem[{Helber et~al.(2019)Helber, Bischke, Dengel, and Borth}]{Eurosat}
Helber, P.; Bischke, B.; Dengel, A.; and Borth, D. 2019.
\newblock EuroSAT: {A} Novel Dataset and Deep Learning Benchmark for Land Use
  and Land Cover Classification.
\newblock \emph{{IEEE} J. Sel. Top. Appl. Earth Obs. Remote. Sens.}

\bibitem[{Jia et~al.(2021)Jia, Yang, Xia, Chen, Parekh, Pham, Le, Sung, Li, and
  Duerig}]{ALIGN}
Jia, C.; Yang, Y.; Xia, Y.; Chen, Y.; Parekh, Z.; Pham, H.; Le, Q.~V.; Sung,
  Y.; Li, Z.; and Duerig, T. 2021.
\newblock Scaling Up Visual and Vision-Language Representation Learning With
  Noisy Text Supervision.
\newblock In \emph{Proc. of ICML}.

\bibitem[{Jia et~al.(2022)Jia, Tang, Chen, Cardie, Belongie, Hariharan, and
  Lim}]{VPT}
Jia, M.; Tang, L.; Chen, B.; Cardie, C.; Belongie, S.~J.; Hariharan, B.; and
  Lim, S. 2022.
\newblock Visual Prompt Tuning.
\newblock In \emph{Proc. of ECCV}.

\bibitem[{Khattak et~al.(2022)Khattak, Rasheed, Maaz, Khan, and Khan}]{Maple}
Khattak, M.~U.; Rasheed, H.~A.; Maaz, M.; Khan, S.; and Khan, F.~S. 2022.
\newblock MaPLe: Multi-modal Prompt Learning.
\newblock \emph{CoRR}.

\bibitem[{Kirkpatrick et~al.(2017)Kirkpatrick, Pascanu, Rabinowitz, Veness,
  Desjardins, Rusu, Milan, Quan, Ramalho, Grabska-Barwinska et~al.}]{WCLoss}
Kirkpatrick, J.; Pascanu, R.; Rabinowitz, N.; Veness, J.; Desjardins, G.; Rusu,
  A.~A.; Milan, K.; Quan, J.; Ramalho, T.; Grabska-Barwinska, A.; et~al. 2017.
\newblock Overcoming catastrophic forgetting in neural networks.
\newblock \emph{Proceedings of the national academy of sciences}.

\bibitem[{Krause et~al.(2013)Krause, Stark, Deng, and
  Fei{-}Fei}]{Stanford_cars}
Krause, J.; Stark, M.; Deng, J.; and Fei{-}Fei, L. 2013.
\newblock 3D Object Representations for Fine-Grained Categorization.
\newblock In \emph{Proc. of ICCV}.

\bibitem[{Lester, Al{-}Rfou, and Constant(2021)}]{PromptTuning}
Lester, B.; Al{-}Rfou, R.; and Constant, N. 2021.
\newblock The Power of Scale for Parameter-Efficient Prompt Tuning.
\newblock In \emph{Proc. of EMNLP}.

\bibitem[{Li et~al.(2022)Li, Weinberger, Belongie, Koltun, and
  Ranftl}]{Segmentation1}
Li, B.; Weinberger, K.~Q.; Belongie, S.~J.; Koltun, V.; and Ranftl, R. 2022.
\newblock Language-driven Semantic Segmentation.
\newblock In \emph{Proc. of ICLR}.

\bibitem[{Liu et~al.(2021)Liu, Yuan, Fu, Jiang, Hayashi, and
  Neubig}]{liu2021pre}
Liu, P.; Yuan, W.; Fu, J.; Jiang, Z.; Hayashi, H.; and Neubig, G. 2021.
\newblock Pre-train, prompt, and predict: A systematic survey of prompting
  methods in natural language processing.
\newblock \emph{arXiv preprint arXiv:2107.13586}.

\bibitem[{Maji et~al.(2013)Maji, Rahtu, Kannala, Blaschko, and
  Vedaldi}]{FGVC_aircraft}
Maji, S.; Rahtu, E.; Kannala, J.; Blaschko, M.~B.; and Vedaldi, A. 2013.
\newblock Fine-Grained Visual Classification of Aircraft.
\newblock \emph{CoRR}.

\bibitem[{Nilsback and Zisserman(2008)}]{Oxford_flowers}
Nilsback, M.; and Zisserman, A. 2008.
\newblock Automated Flower Classification over a Large Number of Classes.
\newblock In \emph{Sixth Indian Conference on Computer Vision, Graphics and
  Image Processing, ICVGIP 2008, Bhubaneswar, India, 16-19 December 2008}.

\bibitem[{Parkhi et~al.(2012)Parkhi, Vedaldi, Zisserman, and
  Jawahar}]{Oxford_pets}
Parkhi, O.~M.; Vedaldi, A.; Zisserman, A.; and Jawahar, C.~V. 2012.
\newblock Cats and dogs.
\newblock In \emph{Proc. of CVPR}.

\bibitem[{Radford et~al.(2021)Radford, Kim, Hallacy, Ramesh, Goh, Agarwal,
  Sastry, Askell, Mishkin, Clark, Krueger, and Sutskever}]{CLIP}
Radford, A.; Kim, J.~W.; Hallacy, C.; Ramesh, A.; Goh, G.; Agarwal, S.; Sastry,
  G.; Askell, A.; Mishkin, P.; Clark, J.; Krueger, G.; and Sutskever, I. 2021.
\newblock Learning Transferable Visual Models From Natural Language
  Supervision.
\newblock In \emph{Proc. of ICML}.

\bibitem[{Recht et~al.(2019)Recht, Roelofs, Schmidt, and Shankar}]{imagenetv2}
Recht, B.; Roelofs, R.; Schmidt, L.; and Shankar, V. 2019.
\newblock Do imagenet classifiers generalize to imagenet?
\newblock In \emph{Proc. of ICML}.

\bibitem[{Schick and Sch{\"u}tze(2021)}]{ManualVerbalizer}
Schick, T.; and Sch{\"u}tze, H. 2021.
\newblock Exploiting Cloze-Questions for Few-Shot Text Classification and
  Natural Language Inference.
\newblock In \emph{Proc. of EACL}.

\bibitem[{Soomro, Zamir, and Shah(2012)}]{UCF101}
Soomro, K.; Zamir, A.~R.; and Shah, M. 2012.
\newblock {UCF101:} {A} Dataset of 101 Human Actions Classes From Videos in The
  Wild.
\newblock \emph{CoRR}.

\bibitem[{Sung, Cho, and Bansal(2022)}]{VQA1}
Sung, Y.; Cho, J.; and Bansal, M. 2022.
\newblock {VL-ADAPTER:} Parameter-Efficient Transfer Learning for
  Vision-and-Language Tasks.
\newblock In \emph{Proc. of CVPR}.

\bibitem[{Wang et~al.(2019)Wang, Ge, Lipton, and Xing}]{imagenet-sketch}
Wang, H.; Ge, S.; Lipton, Z.; and Xing, E.~P. 2019.
\newblock Learning robust global representations by penalizing local predictive
  power.
\newblock \emph{Proc. of NeurIPS}.

\bibitem[{Xiao et~al.(2010)Xiao, Hays, Ehinger, Oliva, and Torralba}]{SUN397}
Xiao, J.; Hays, J.; Ehinger, K.~A.; Oliva, A.; and Torralba, A. 2010.
\newblock {SUN} database: Large-scale scene recognition from abbey to zoo.
\newblock In \emph{Proc. of CVPR}.

\bibitem[{Yao et~al.(2022)Yao, Huang, Hou, Lu, Niu, Xu, Liang, Li, Jiang, and
  Xu}]{FILIP}
Yao, L.; Huang, R.; Hou, L.; Lu, G.; Niu, M.; Xu, H.; Liang, X.; Li, Z.; Jiang,
  X.; and Xu, C. 2022.
\newblock {FILIP:} Fine-grained Interactive Language-Image Pre-Training.
\newblock In \emph{Proc. of ICLR}.

\bibitem[{Zang et~al.(2022)Zang, Li, Zhou, Huang, and Loy}]{Detection2}
Zang, Y.; Li, W.; Zhou, K.; Huang, C.; and Loy, C.~C. 2022.
\newblock Open-Vocabulary {DETR} with Conditional Matching.
\newblock In \emph{Proc. of ECCV}.

\bibitem[{Zhai et~al.(2022)Zhai, Wang, Mustafa, Steiner, Keysers, Kolesnikov,
  and Beyer}]{LIT}
Zhai, X.; Wang, X.; Mustafa, B.; Steiner, A.; Keysers, D.; Kolesnikov, A.; and
  Beyer, L. 2022.
\newblock LiT: Zero-Shot Transfer with Locked-image text Tuning.
\newblock In \emph{Proc. of CVPR}.

\bibitem[{Zhou et~al.(2022{\natexlab{a}})Zhou, Yang, Loy, and Liu}]{CoCoOp}
Zhou, K.; Yang, J.; Loy, C.~C.; and Liu, Z. 2022{\natexlab{a}}.
\newblock Conditional Prompt Learning for Vision-Language Models.
\newblock In \emph{Proc. of CVPR}.

\bibitem[{Zhou et~al.(2022{\natexlab{b}})Zhou, Yang, Loy, and Liu}]{CoOp}
Zhou, K.; Yang, J.; Loy, C.~C.; and Liu, Z. 2022{\natexlab{b}}.
\newblock Learning to Prompt for Vision-Language Models.
\newblock \emph{Int. J. Comput. Vis.}

\end{thebibliography}

\end{document}